# Design, Modelling and Experimental Evaluation of a Tendon-driven Wrist Abduction-Adduction Mechanism for an upper limb exoskeleton

Juwairiya S. Khan[1,2], Mostafa Mohammadi[1], John Rasmussen[2], and Lotte N.S. Andreasen Struijk[1]

***Abstract*—Wrist exoskeletons play a vital role in rehabilitation and assistive applications, yet conventional actuation mechanisms such as electric motors or pneumatics often introduce undesirable weight, friction, and complexity. This paper presents a novel single-cable (tendon), torsional-spring-assisted actuation mechanism for wrist abduction-adduction, and a simulation-based method for selecting its stiffness parameters. The mechanism employs a single Bowden cable passively tensioned by a spiral torsional spring (clock spring) to maintain continuous cable tension without antagonistic actuation. Kinematic and dynamic modeling of the mechanism was performed to estimate the required torque and identify optimal spring parameters. These simulation-derived parameters guided the design of a functional prototype, which was experimentally evaluated with five participants with no motor disabilities (NMD) under varying arm positions and loading conditions using three spring configurations to account for user variability and modeling uncertainties. Experimental results show consistent agreement with simulation-derived trends, with the nominal spring configuration achieving balanced motion range, torque demand, and repeatability. The results demonstrate that simulation-informed stiffness selection can effectively guide the design of compact, cable-driven wrist exoskeletons while reducing reliance on empirical tuning.**

***Index Terms*— Wrist Abduction-Adduction, Exoskeleton, cable-driven mechanism, Assistive rehabilitation, experimental validation**

## I. Introduction

Upper limb motor impairments caused by neuromusculoskeletal disorders such as spinal cord injury (SCI) and amyotrophic lateral sclerosis (ALS), can severely limit range of motion (ROM) and the ability to perform activities of daily living (ADLs) [1], [2]. Upper limb exoskeletons (ULEs) have emerged as a promising assistive technology to restore partial autonomy by augmenting neuromuscular function while preserving the user's limb[3], [4] [5].

A central challenge in exoskeleton design is reducing device mass, inertia and volume [6], which can contribute to discomfort, joint misalignment, and fatigue [7]. This problem is most acute in distal joints such as the wrist, where additional mass and bulky objects attached, can severely degrade natural movement [8], [9], [10]. Cable-driven transmissions address this by relocating actuators proximally, reducing distal inertia[11], [12][13]. However, cable-driven systems introduce slack, nonlinear tension, friction, and elastic deformation degrade transparency, torque fidelity, and control accuracy [14]. These limitations are particularly problematic for bidirectional motions such as wrist abduction-adduction (Ab-Ad), a degree of freedom often neglected in existing exoskeletons [15].

Existing solutions to mitigate slack include antagonistic cable pairs, preloaded tensioners, and active tension control [16]. While effective, they increase mechanical complexity, weight, and routing requirements, counteracting the benefits of cable transmission [13], [17]. Passive elastic elements have been explored to improve efficiency and reduce actuator demand [18], but these approaches largely focus on actuator-level design. Biomechanics-driven specification of passive stiffness for exoskeleton applications remains limited. Thus, there is a critical need for a low-profile solution that combines the lightweight benefits of cable transmission with robust tension management.

To the best of our knowledge, active bidirectional tendon-driven actuation realized by a single cable and a spiral torsional spring (clock spring) has not been systematically investigated for wrist Ab-Ad in rehabilitation or assistive contexts. Wrist Ab-Ad, together with flexion-extension are necessary for performing ADLs and previous designs either focus primarily on flexion-extension [19] or rely on multi-cable and multi-motor systems with increased weight and complexity [20], [21].

This paper presents the design, modeling, and experimental validation of a novel actuation mechanism for wrist Ab-Ad exoskeleton (WristExo) joint that addresses these specific limitations. The design integrates a preloaded clock spring with a Bowden cable (tendon) to maintain continuous tension, thereby eliminating the need for antagonistic tendon pairs or active tension control. This configuration not only reduces mechanical complexity and bulk but also enables passive energy

*Research supported by Aage og Johanne Louis-Hansens Fond.

[1]Juwairiya Siraj Khan (corresponding author), Mostafa Mohammadi and Lotte N.S. Andreasen. Struijk are with the Neurorehabilitation Robotics and Engineering, Center for Rehabilitation Robotics, Department of Health Science and Technology, Aalborg University, Aalborg East, Denmark. (jsikh@hst.aau.dk , mostafa@hst.aau.dk, naja@hst.aau.dk ).

[2]Juwairiya Siraj Khan and John Rasmussen are with the Department of Materials and Production, Aalborg University, Aalborg East, Denmark (e-mail: jsikh@mp.aau.dk, jr@mp.aau.dk)

storage and release. Lightweight 3D-printed components and a compact joint architecture further enhance wearability and system responsiveness, suitable for integration into a full upper-limb exoskeleton [22].The design process was guided by kinematic and dynamic modeling in the AnyBody Modeling System (AMS) [23], which quantified wrist torque requirements and optimized the spring's stiffness and pretension.

This modeling-based, user-centered approach reduced design iterations and informed prototype development [24]. A lightweight 3D-printed exoskeleton was fabricated and experimentally evaluated with five participants with no motor disability (NMD) across multiple arm positions and loading conditions. Three spring variants were tested, showing strong agreement between simulation and experimental results, confirming the system's ability to deliver consistent assistive torque and smooth motion. Compared to antagonistic cable configurations, the proposed single-cable architecture reduces actuator count, routing complexity, and control requirements, while maintaining continuous tension through passive elasticity. This results in lower mechanical complexity and reduced distal mass, without requiring active tension control. The main contributions of this paper are threefold:

1. Mechanism Design: A single-cable, clock-spring-assisted wrist Ab-Ad actuation architecture that eliminates antagonistic actuation while maintaining continuous cable tension. minimizing mechanical complexity.

2. Modeling: A biomechanics-driven approach using the AMS to estimate wrist torque requirements and derive corresponding spring stiffness and pretension parameters.

3. Evaluation: Experimental validation with five participants across multiple arm positions and loading conditions, demonstrating consistent agreement between simulation predictions and experimental performance.

The remainder of this paper is structured as follows: Section II details the mechanism working principle and its implementation. Section III describes the kinematic and dynamic modeling approach. Section IV outlines the experimental protocol. Section V and VI presents and discuss the results respectively. Finally, Section VI offers concluding remarks and suggests directions for future research.

## II. MECHANISM DESIGN AND IMPLEMENTATION

The design of the wrist actuation mechanism was driven by the need to provide adequate biomechanical support for Activities of Daily Living (ADLs) while prioritizing user comfort, a low profile, and minimal perceived weight. This section details the requirements, conceptual principle, mechanical realization, and critical component selection for the proposed cable-spring mechanism.

### *A. Biomechanical Requirements of Wrist Abduction-Adduction*

The design specifications were derived from anthropometric data, biomechanical studies and end-user inputs during user-board meetings of users with SCI and ALS [24]. This was done to ensure the exoskeleton can accommodate a wide user population and generate sufficient assistive torque with maximum comfort and usability.

1. Hand Sizes and Ergonomics: The interface was designed for the 5$^{th}$-95th percentile of adult hand dimensions based on anthropometric data in literature. Mean hand length is 197.1 ± 9.3 mm (male) and 179.3 ± 8.6 mm (female), and breadth is 89.7 ± 4.1 mm (male) and 76.9 ± 3.8 mm (female) [25]. Hand segment weight corresponds to 0.65 ± 0.06% (male) and 0.50 ± 0.026% (female) of body weight [26], with scaling applicable across populations [27], [28]. A worst case (large hand) defined maximum dimensions, while adjustable straps and padding ensured fit for smaller users; three brace sizes (small, medium, large) were implemented.

2. Range of Motion (ROM)**:** Mean ROM for wrist abduction-adduction joint from literature is 81±16° (abduction: 24±9° adduction: 57±9°) for males and 79±17° (abduction: 24±11° adduction: 55±11°) for females [29], [30]. However, functional ADLs require approximately ±10-30° of Ab-Ad movement [31], [32]. Therefore, the mechanism was designed to accommodate a conservative ROM of 30° abduction and 44° adduction to cover all functional tasks required for ADLs.

3. Torque and Current Requirements: Biomechanical modeling and literature review indicate that a maximum torque of 0.5 Nm is sufficient to assist with most ADLs [33]. This torque value became the primary design target. The current requirements for the DC motor were subsequently sized to generate this torque at the joint, accounting for transmission efficiency.

4. Comfort: A paramount requirement was to minimize the device's weight and inertia on the wrist. The target moving mass of the wrist module was set to < 250 g to achieve a lightweight wrist exoskeleton, to be later integrated into a whole arm exoskeleton. All edges were rounded, and interfaces were padded to distribute pressure evenly and avoid soft tissue compression during movement.

### *B. Concept and Working Principle*

The proposed mechanism employs a hybrid actuation strategy combining an active Bowden cable with a passive clock spring to enable smooth, bidirectional control of wrist Ab-Ad. The working principle is illustrated in Fig.1. The core innovation is the exploitation of the clock spring at the joint as an antagonist in a tendon-driven wrist joint.

1. Activation (Wrist Abduction/Wrist neutral driven by Bowden cable): To initiate movement, the DC motor winds the cable anchored at the shaft. This pull rotates the shaft connected to exoskeleton's hand frame, abducting the wrist (Fig. 1 (a)).

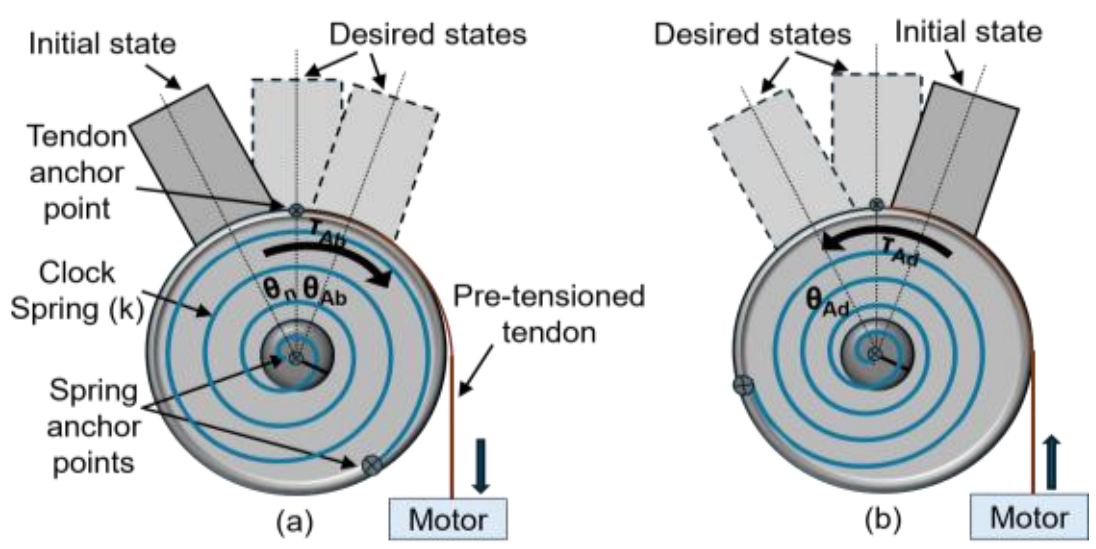

Fig. 1 Tendon-driven joint mechanism with clock spring (a) joint moves from adducted to neutral/abducted state (b) returns to neutral/adducted state

Simultaneously, this winding action also rotates the spool housing the clock spring, thereby winding and storing potential energy in the spring.

2. Return (Adduction/Wrist neutral driven by spring): To reverse the motion, the DC motor is simply reversed or disengaged. The energy stored in the wound clock spring is then released, providing a restorative torque that unwinds the cable and passively adducts the joint to the desired position as shown in Fig.1(b).

This mechanism maintains continuous cable tension throughout the motion range due to spring pretension, reducing slack and backlash effects. This is enabled by the pretension of the clock spring, which minimizes the torque variation over the range of motion.

### C. *Mechanical Design*

The mechanical design (Solidworks 2025) of WristExo was realized through a combination of 3D-printed (using Flashforge Creator 4) polymers and off-the-shelf components to achieve a lightweight and cost-effective prototype shown in Fig. 2. The design emphasizes compactness, low friction, modularity, lightweight construction, and anatomical alignment to ensure both functional performance and user comfort.

1. Hand brace and Joint: The structure is 3D-printed from polylactic acid (PLA) and PLA carbon fiber (PLA-CF) to achieve a high strength-to-weight ratio. It incorporates an adjustable single-axis joint aligned with the anatomical wrist Ab-Ad axis. The dorsal hand brace is secured using adjustable Velcro straps across the metacarpals.
2. Base module: The base module integrates the cable transmission and clock spring (Lesjofors Denmark) assembly, with its rotational axis aligned to the wrist Ab-Ad axis. Spring's inner end is anchored to this axis to generate recoil torque. A tubular guide routes the Bowden cable tangentially to optimize torque transmission. It also incorporates needle roller bearings to reduce friction, and the module is secured to the forearm via adjustable straps.
3. Clock Spring Housing: The housing fixes the spring's outer end and rotates under cable actuation, storing energy during abduction and releasing it for passive adduction. It also serves as a mechanical interface to the hand brace. The housing functions as a lever arm, converting cable tension into joint torque as in (1):

$$\boldsymbol{\tau} = \mathbf{r} \times \mathbf{F} \quad (1)$$

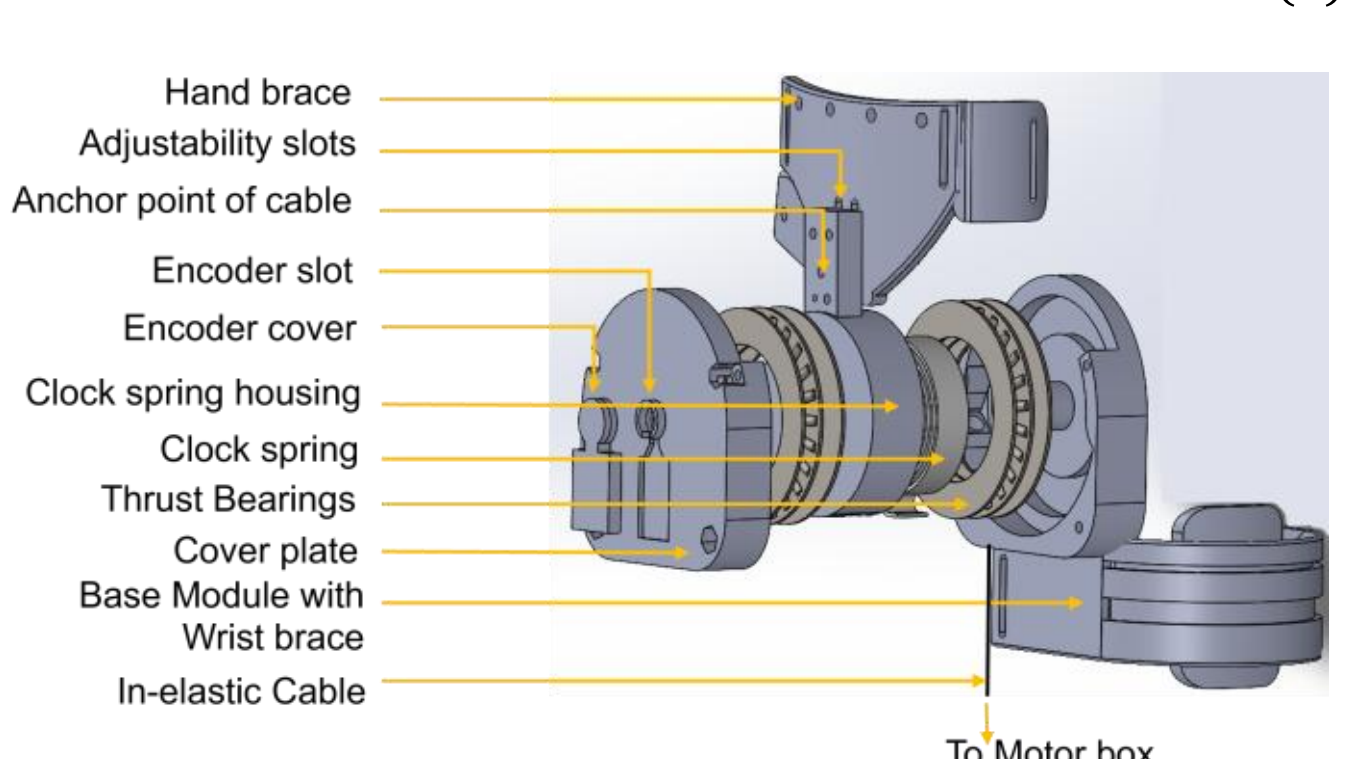

Fig. 2 Exploded view of CAD assembly schematic of the WristExo

where $\boldsymbol{\tau}$ is joint torque, $\mathbf{F}$ is cable tension vector and $\mathbf{r}$ is the radius vector from joint axis to cable attachment (lever arm)

4. Cable Transmission: A 1 mm diameter, high-strength, low-stretch braided steel cable (Special-Butikken, Ribe A/S) runs from a remote motor capstan through a low-friction Bowden sheath into the base module. The cable wraps tangentially around the clock spring housing to generate abduction torque. Routing and attachment geometry were optimized to minimize backlash and maximize torque efficiency.
5. Cover Plate and Sensing: A cover plate maintains structural alignment and integrates an absolute magnetic encoder (RM08, RLS) for joint angle measurement. Internal needle bearings are used to further reduce friction.
6. Actuation Unit: The actuation unit employs a DC motor assembly (Maxon EC-4 pole 22 mm) with planetary gearhead of 128:1 reduction (Maxon GP 22 mm) fixed to the motor-box base, with its output shaft directly coupled to a winding spool. The cable is coiled on the spool and transmitted to the joint through a Bowden tube, enabling remote actuation. Both the motor box and controller (Maxon EPOS4 50/8 CAN) are positioned proximally as demonstrated in Fig 3 (a).

### D. **Spring Selection and Pretensioning**

The clock spring is the critical passive element defining the system's torque-characteristic and return capability. Its selection was based on kinematic and dynamic modeling.

1. Spring Selection: The required spring stiffness, 12.32 N·mm/deg, was derived from the target assistive torque profile obtained through AMS modeling. Simulations indicated an approximately constant torque demand across the wrist's range of motion, making a constant-force clock spring the most suitable choice. To accommodate inter-user variability in joint stiffness and comfort, three off-the-shelf clock springs ($k_1$ = 10.66, $k_2$ = 11.71, and $k_3$ = 13.2 N·mm/deg) were selected for experimental validation.
2. Pre-tensioning: Spring pretension ensures return motion from the neutral position and defines the rest pose *(0°)*. During assembly, the spring is wound by a predetermined number of turns *($\theta_{pre}$)* before being secured in its housing. This pre-winding establishes an initial baseline torque $\tau_{pre} = k * \theta_{pre}$. This pretension was optimized in AMS to exceed static friction while maintaining user comfort at rest. When selecting the clock springs to match the simulated parameters with off-the shelf springs, mounting of spring (S1) caused pretension variability.

## III. Modeling of the Actuation system

### A. *Kinematic and Dynamic Modeling (AMS)*

The exoskeleton-arm system was modeled using AMS to simulate wrist Ab-Ad assistance during three different arm positions, including resting, reaching, and drinking-like postures. The arm was represented by upper arm, forearm, and hand segments connected through anatomical joints, while the exoskeleton comprised aligned revolute joints at the wrist. The model included segmental mass properties and incorporated the mass of the handheld load (0.5 kg) as well as the weight of the hand and forearm, considering the worst case of hand sizes and weights as mentioned in section II A [26], [27], [28].

Three representative arm configurations were analyzed to capture the ROM during assisted tasks: The shoulder and elbow were kinematically driven to represent the three positions, Position 1: neutral resting posture (shoulder flexion ≈ 30°, elbow flexion ≈ 60°, forearm pronation ≈ 90°), position 2: forward-reaching posture (shoulder flexion ≈ 45°, elbow flexion ≈ 60°, forearm at neutral position), and position 3: drinking posture (shoulder flexion ≈ 75°, elbow flexion ≈ 120°) as demonstrated in Fig. 3(a-c), while the wrist joint motion was defined using a Fourier driver to replicate one cycle of Ab-Ad. The exoskeleton's assistive torque was generated through a modeled spiral torsional spring acting about the wrist joint. The cable path was modeled using a spline-based wrapping constraint which enabled precise computation of the cable geometry.

### B. Cable Tension and Spring Torque Estimation

The exoskeleton's wrist joint is actuated by a clock spring mechanism that delivers torque through a Bowden cable. The spring torque $T_s$ is defined as in (2):

$$T_s = k(\theta + \theta_0) \quad (2)$$

where $k$ is the spring constant *[Nm/rad]*, $\theta$ *[rad]* is the angular displacement of the wrist joint, and $\theta_0$ is the pretension angle representing the initial preload. These two parameters directly determine the magnitude and smoothness of the assistive torque. Torque computed from motor current represents motor-side torque while joint torque is obtained through gear reduction and cable transmission.

To determine the nominal spring parameters, the reaction moment versus joint angle data from all three positions were modelled and the worst-case position (maximum torque) was fitted with a linear trendline using least-squares regression [34]. The slope of the fitted line represents effective stiffness $k$ as in (3), while x-intercept yields the pretension $\theta_0$ as in (4):

$$k = \frac{dT_{sim}}{d\theta} \quad (3)$$

$$\theta_0 = -\frac{intercept}{k} \quad (4)$$

This fit provides a nominal spring configuration that offers comparative torque assistance across representative ADLs.

### C. Friction Modeling

Cable-sheath friction was included in the model using the capstan equation as in (5):

$$F_{out} = F_{in} e^{\mu \theta_w} \quad (5)$$

where $F_{in}$ and $F_{out}$ denote the input and output cable forces, μ is the friction coefficient, and $\theta_w$ is the total wrap (bend) angle of the cable from the motor box to the anchor point in the capstan. A friction coefficient of 0.04 was adopted based on prior studies of Bowden cable actuation. This formulation accounts for the nonlinear reduction in transmitted force, ensuring realistic torque predictions in exoskeleton-assisted motion.

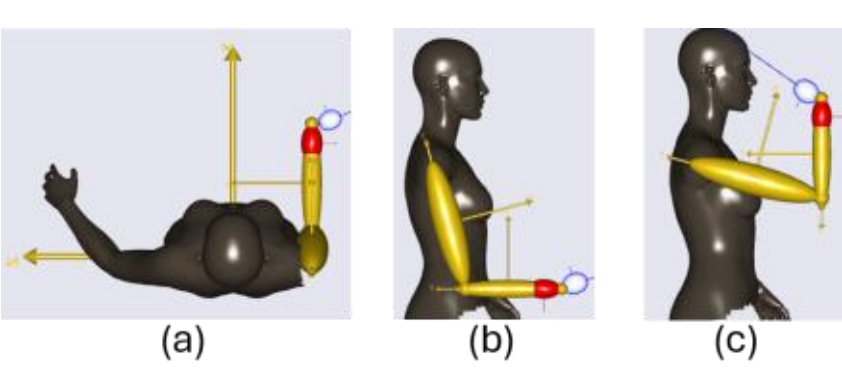

Fig. 3 Joint mechanism Modeling in different positions in AMS (a) Arm Position 1, (b) Arm Position 2, (c) Arm Position 3

## IV. EXPERIMENTAL VALIDATION

### A. Participants

Five able bodied adults (4 male, 1 female; age range: 25-35 years) participated in the study. All participants signed a written informed consent form, and the experimental protocol was approved by the regional ethics committee (The Scientific Ethics Committee for the North Jutland region, VEK, Protocol number: 20220029). Participants had no history of musculoskeletal or neurological disorders affecting wrist function. Out of five, only one participant (P2) was left-handed.

### B. Experiment Setup and Procedure

The wrist exoskeleton was mounted on the participant's right hand as demonstrated in Fig. 4. The experimental protocol evaluated exoskeleton performance across multiple arm positions and loading conditions representative of ADLs (e.g., drinking) as shown in Fig 4 (b-d). Hand measurements and wrist ROM (using digital goniometer) were recorded prior to testing to adapt the device to each participant [30]. Participants performed wrist Ab-Ad under three arm positions (Section III-A) and two load conditions (unloaded and 300 g). Three clock springs were tested in randomized order across all conditions, representing nominal stiffness $k$ and two variants $(k + 1, k - 1)$ derived from AMS simulations to capture design tolerances and user variability [35]. For each condition (position × load × spring), two motion sequences were performed, each consisting of abduction (triggered by Button 2), adduction (Button 3), and return to neutral (Button 4) via gamepad (Thrustmaster Dual Analog 5) control. The system operates in position-controlled mode, where the motor drives the cable while the clock spring provides passive return torque. A training trial preceded data collection. Each session lasted approximately 3 hours, including 15 min rest periods to mitigate fatigue. The experiment timing is shown in Fig. 5. Control and data acquisition were implemented in ROS, with motor control in C++ and user interface modules in Python.

### C. Data Acquisition

Sensor data were recorded using the Robot Operating System (ROS) Noetic at 100 Hz. The following signals were logged:

1. Joint angle: measured via a magnetic encoder (RLS RM08) mounted on the exoskeleton joint.
2. Motor current: sampled through a digital current sensor, converted to torque using the motor's torque constant $K_t$.
3. Gamepad commands: button presses corresponding to wrist abduction from neutral (Button 2), wrist adduction from abducted position (Button 3), and return to neutral (Button 4).
4. Timestamps: synchronized across data streams via ROS time.

### D. Data Processing

Raw data were processed using a custom Python script. Missing or implausible data points (<5% of the dataset) were handled using linear interpolation based on adjacent valid samples within the same trial. This approach was applied only to isolated instances and did not affect overall trends or statistical outcomes.

1. Kinematic and Kinetic Metrics

From the joint angle data, the wrist abduction and adduction ranges and the total joint ROM were computed as in (6):

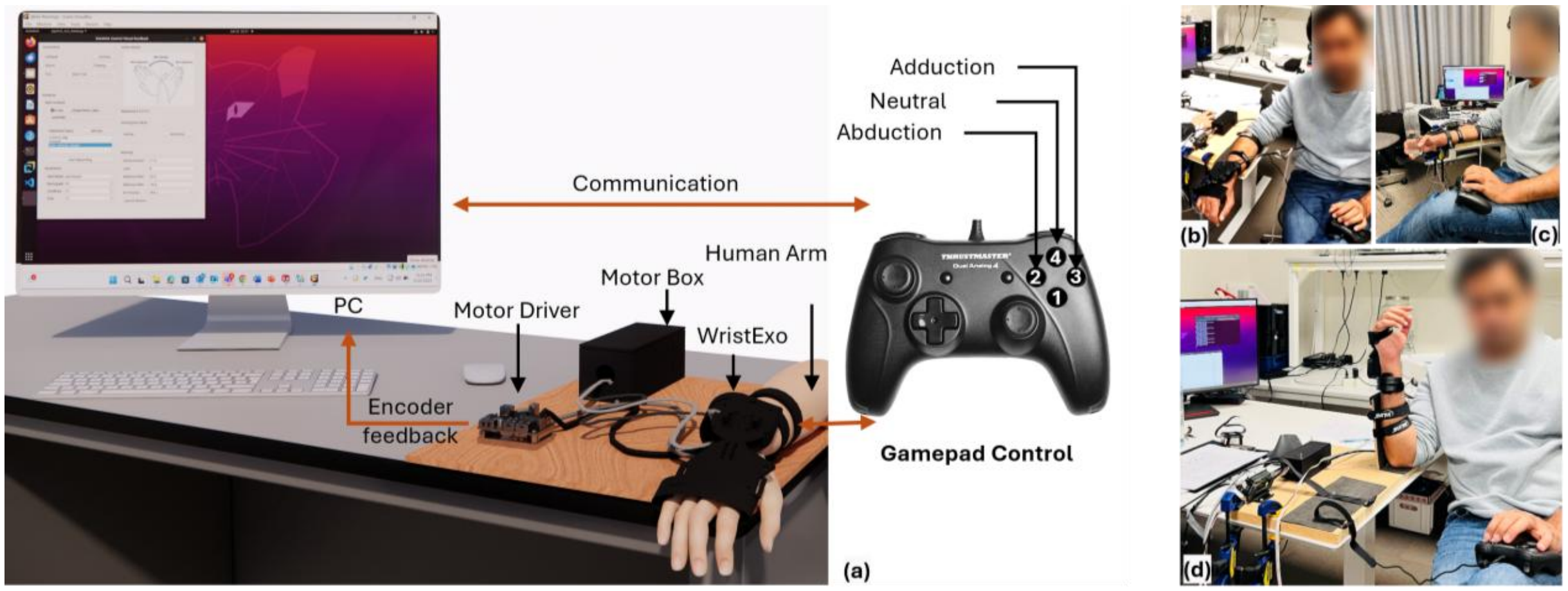

Fig. 4 Experimental setup overview (a) Demonstration of WristExo setup and gamepad control and snapshots from the experimental evaluation with a participant in different positions (b) Position 1, (c) Position 2, (d) Position 3

$$ROM_{Ab} = max(\theta(t)), ROM_{Ad} = | \, min(\theta(t)) \, |, \quad ROM_{total} = ROM_{Ab} + ROM_{Ad} \tag{6}$$

where θ(t) is the angular position of the wrist joint over time.

Motor torque was estimated from the motor current *I(t)* using the actuator torque constant $K_T$=0.0105 Nm/A, as in (7):

$$\tau(t) = K_T \cdot I(t) \tag{7}$$

The measured motor current (in mA) was converted to Amperes prior to torque computation using the motor torque constant $K_T$ and the root-mean-square (RMS) torque was used to characterize actuator demand, as it reflects the effective continuous load during motion and is less sensitive to transient peak variations. It was obtained for each trial as in (8):

$$\tau_{RMS} = K_T \sqrt{\frac{1}{N}\left(\sum_{i=1}^{N} I_i^2\right)} \tag{8}$$

2. Statistical Analysis

Non-parametric statistical analysis was used to compare spring performance. The Friedman test evaluated within-subject differences among the three springs (S1-S3) for ROM and RMS torque. Repeatability was quantified as the absolute difference between repeated trials ($|T_1 - T_2|$). Box plots were used to summarize ROM and torque distributions across spring configurations. Repeatability was analyzed separately as a function of spring and wrist position. All analyses and visualizations were implemented in Python using Seaborn, Matplotlib, and SciPy.

### E. *Usability Questionnaire*

Following the experimental session, participants completed a brief usability questionnaire comprising three items assessing perceived device size, weight, and ease of donning/doffing. Responses were recorded on a 10-point Likert scale (1 = Not likely, 10 = Unacceptably likely), where higher scores indicated greater perceived difficulty or discomfort. The questionnaire was designed to provide a preliminary assessment of perceived ergonomics and usability.

## V. Results

This section presents the simulation and experimental evaluation of the proposed wrist exoskeleton across different spring configurations and conditions. Simulations for three arm postures, resting, reaching, and drinking were performed to evaluate wrist joint loading during ADLs. Reaction moments from AMS were used to estimate the nominal spring constant and pretension for the exoskeleton, which were then validated experimentally with five participants performing the same wrist motion. Overall, the results demonstrate that simulation-derived stiffness parameters can effectively guide spring selection, reducing reliance on empirical tuning while achieving consistent performance across configurations.

### A. *AMS modeling results:*

The reaction moment versus joint angle plots obtained from AMS simulations exhibited predominantly linear behavior across all three arm configurations (Fig. 3) for both abduction and adduction movements as shown in Fig. 6(a). Arm position 3 is the worst case due to high torque requirement during adduction phase. Therefore, we considered position 3 to derive the optimal spring parameters for design. As shown in Fig 6(b) the regression of torque versus joint angle produced the following trendline as in (9) indicating a linear stiffness response with a slope magnitude of 0.71 Nm/rad and an intercept of 0.4157 Nm. From this relationship, the effective spring constant and pretension angle were calculated as: $k = 0.71\ Nm/rad, \theta_0 = 0.59\ rad$ (≈ 33.8 ∘).

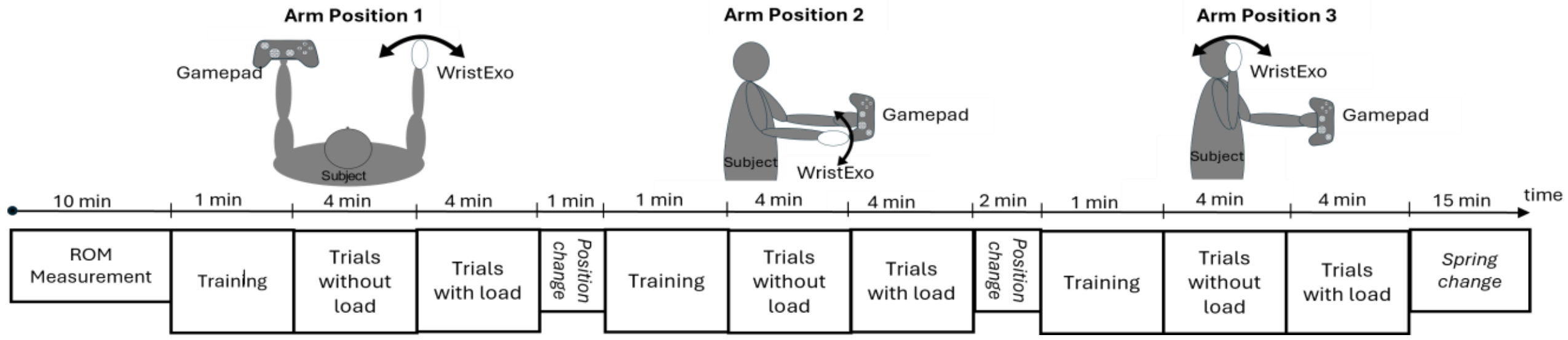

Fig. 5 Experimental Protocol (timing diagram)

$$T_{sim} = -0.7054\theta + 0.4157 \quad (9)$$

The negative slope reflects the defined rotation direction in the model, where spring torque decreases with increasing wrist angle (i.e., the spring unwinds to assist motion). The relatively high linearity ($R^2 \approx 0.97$) indicates consistent behavior of the simulated passive torque across ADL posture. These parameters defined the target stiffness range used to design the evaluated spring configurations (S1-S3), spanning lower, nominal, and higher stiffness values around the simulation-derived optimum.

### B. *Experimental Validation results*

The total wrist range of motion (ROM) across participants was primarily concentrated between 40° and 65°, with some values extending up to ≈78°. Fig. 7 (a) presents boxplots of total ROM, and Fig. 7 (b) shows RMS torque values across the three springs (S1-S3). The medium-stiffness spring (S2) demonstrated the most balanced performance, achieving an average ROM of approximately 50° ± 5° with moderate torque demand (≈ 5 Nmm or 0.005 Nm). The reported RMS torque corresponds to motor-side torque. Considering the 128:1 gear reduction and transmission efficiency, the resulting joint torque (≈0.5 Nm) is consistent with the simulation-derived requirement. Notably, S2 corresponds most closely to the stiffness predicted by the simulation model, supporting the effectiveness of the simulation-driven design approach. In contrast, the softest spring (S1) exhibited larger ROM amplitudes but with increased variability in actuator demand near motion extremes, while the stiffest spring (S3) limited ROM to 40-55° with higher torque variability and peak values.

Friedman's non-parametric tests confirmed that no statistically significant differences were detected across spring configurations, $p > 0.05$ for all ROM ($\chi^2$=2.800, p=0.247) and torque ($\chi^2$=1.200, p=0.549) measures. Nevertheless, the observed mean trends suggest practical differences in user experience and mechanical transparency.

### C. *Repeatability Analysis*

Repeatability, quantified as the absolute difference in ROM between consecutive trials ($|T_1 - T_2|$), is shown in Fig. 8(a). Across all participants, positions, and conditions, S2 exhibited the lowest variability (≈ 2.5-3°), while S1 and S3 showed higher variability (≈ 4-5°), with S3 generally exhibiting the largest spread. Position-wise analysis (POS1-POS3) indicates an increase in variability toward more distal wrist configurations, particularly for S3, suggesting greater sensitivity to alignment and passive mechanical effects. The overall mean (±SD) further confirms that S2 maintains lower variability across conditions without sensitivity to positional changes. This behavior aligns with the simulation-based stiffness selection, where the intermediate stiffness (S2) minimizes variability without introducing excessive constraint, likely due to improved matching with the torque-angle profile and reduced sensitivity to alignment and frictional effects.

### D. *Questionnaire outcomes*

The user evaluation indicated ratings predominantly in the lower range of the adopted Likert scale, where 1 denotes Not likely and 10 denotes Unacceptably likely. Individual participant ratings are shown in Fig.8 (b). The device size scored highest likelihood (Mean = 3.2 and S.D = 2.5) while device weight was rated lower (Mean = 2.2 and S.D = 1.3), indicating acceptable ergonomics. Ease of donning and doffing received the most favorable ratings (Mean = 1.8 and S.D = 1.3), suggesting that the device can be worn and removed with minimal perceived effort.

## VI. DISCUSSION

The simulation and experimental results support the proposed single-cable, spring-assisted mechanism as a consistent solution for wrist Ab-Ad assistance. The strong linearity of the AMS-derived torque-angle relationship ($R^2 \approx 0.97$) enabled direct estimation of stiffness and pretension, which translated to prototype performance. The identified parameters (k = 0.71 Nm/rad, $\theta_0$ = 0.59 rad) define a nominal operating point for ADL-relevant assistance. Inclusion of hand and external loads improved model realism, while the linear torque-angle relationship justifies a clock spring approximation. Compared to multi-cable systems [21] and multi-spring mechanisms [15], the proposed single-cable design provides bidirectional assistance via passive stiffness, reducing complexity and distal mass.

No statistically significant differences were detected between spring configurations, likely due to inter-subject variability, hardware ROM limits, and the small sample size. Variations in effective pretension likely explain the higher ROM and variability observed for spring S1 [36]. From an engineering perspective, spring selection is task-dependent: spring S1 maximizes ROM, while springs S1 and S2 reduce actuator demand relative to spring S3. For the intended application, spring S2 provides the most balanced performance, achieving a near-target ROM with lower torque demand and improved repeatability. Its alignment with simulation predictions supports the validity of the modeling-based design approach.

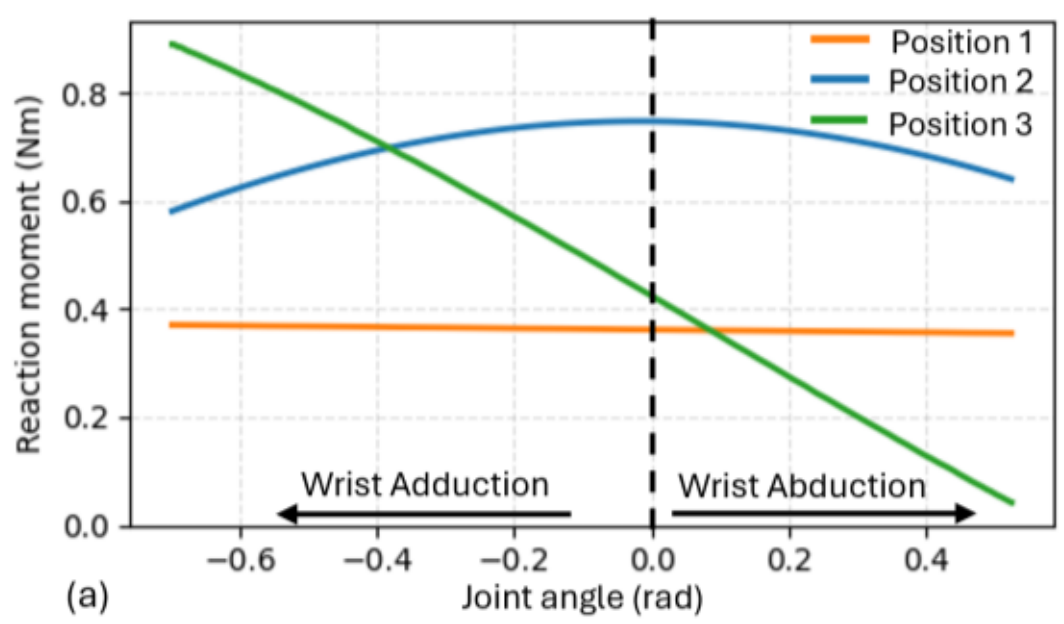

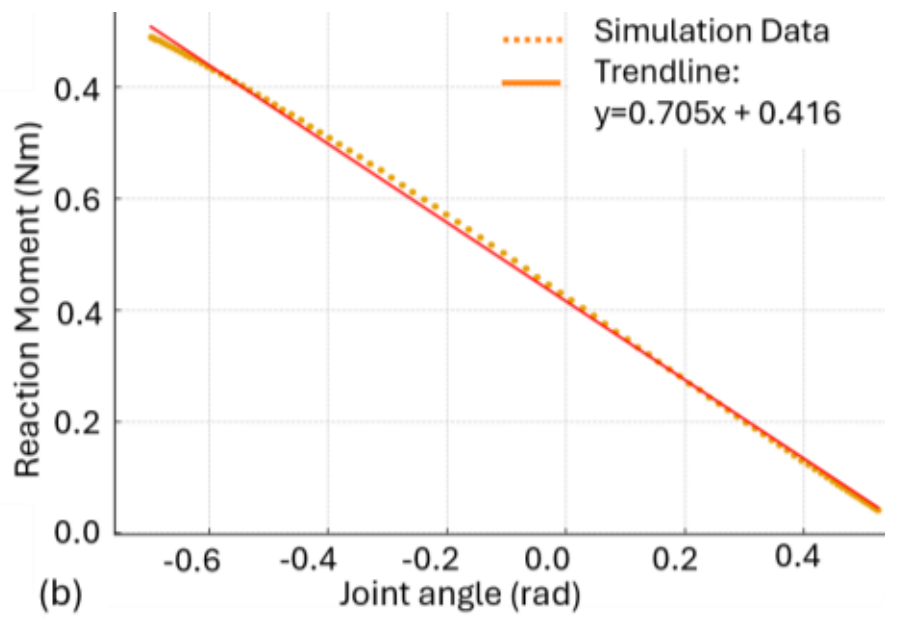

Fig. 6 AMS Simulation results (a) Wrist joint reaction moment vs joint angle in different positions, (b) Joint Reaction moment vs Joint angle relationship for position 3 with a trendline representative of the optimal clock spring

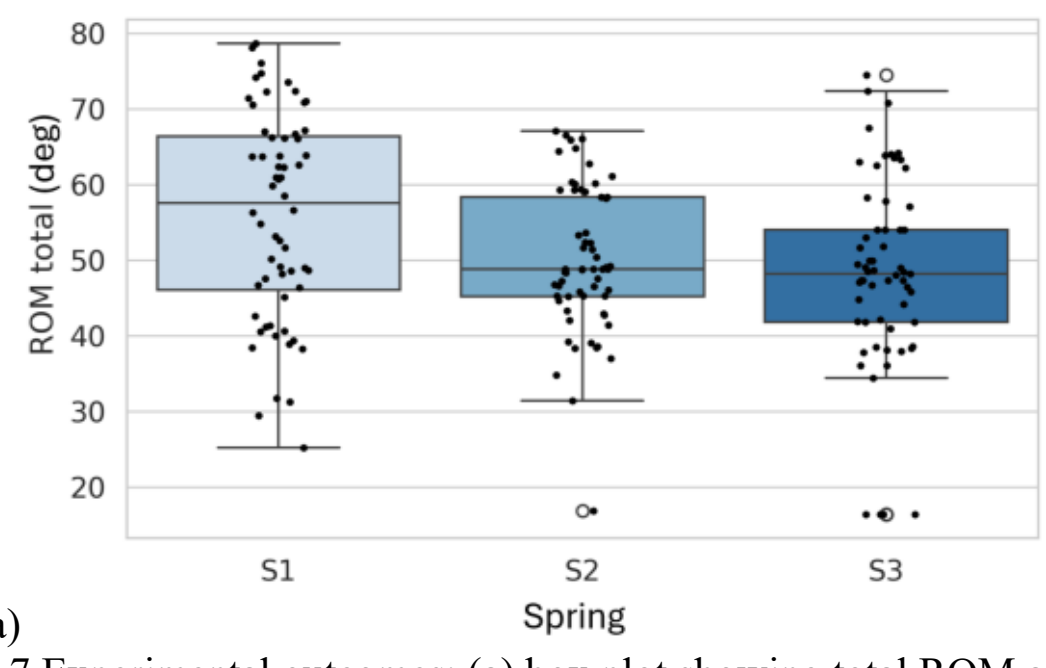

(a)

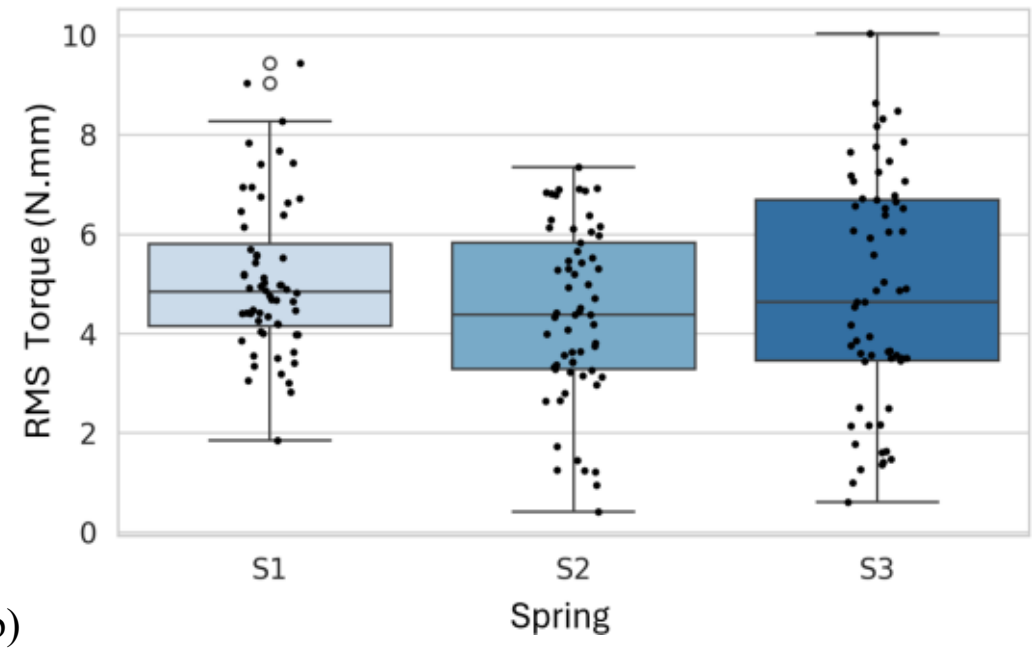

(b)

Fig. 7 Experimental outcomes: (a) box plot showing total ROM and (b) box plot showing RMS torque (in Nmm for readability) for the three spring configurations

This selection is based on practical performance rather than statistical significance. Interpretation is limited by (i) partial data reconstruction for one participant, (ii) hardware-constrained ROM, and (iii) limited statistical power. Despite these constraints, repeatability analysis indicates consistently lower variability for S2, suggesting reduced sensitivity to alignment and passive mechanical effects. As the objective is mechanical validation rather than clinical efficacy, a small cohort is sufficient to assess consistency, torque trends, and repeatability.

User feedback indicates acceptable ergonomics, with low perceived difficulty in donning/doffing and moderate concerns regarding size and weight. Safety limits prevented over-extension with minor ROM exceedances. attributed to measurement and calibration uncertainties. Overall, the results demonstrate that simulation-informed stiffness selection can enable predictable performance in passive cable-driven exoskeletons while reducing reliance on empirical tuning. The proposed mechanism provides a practical basis for future integration and adaptive stiffness control.

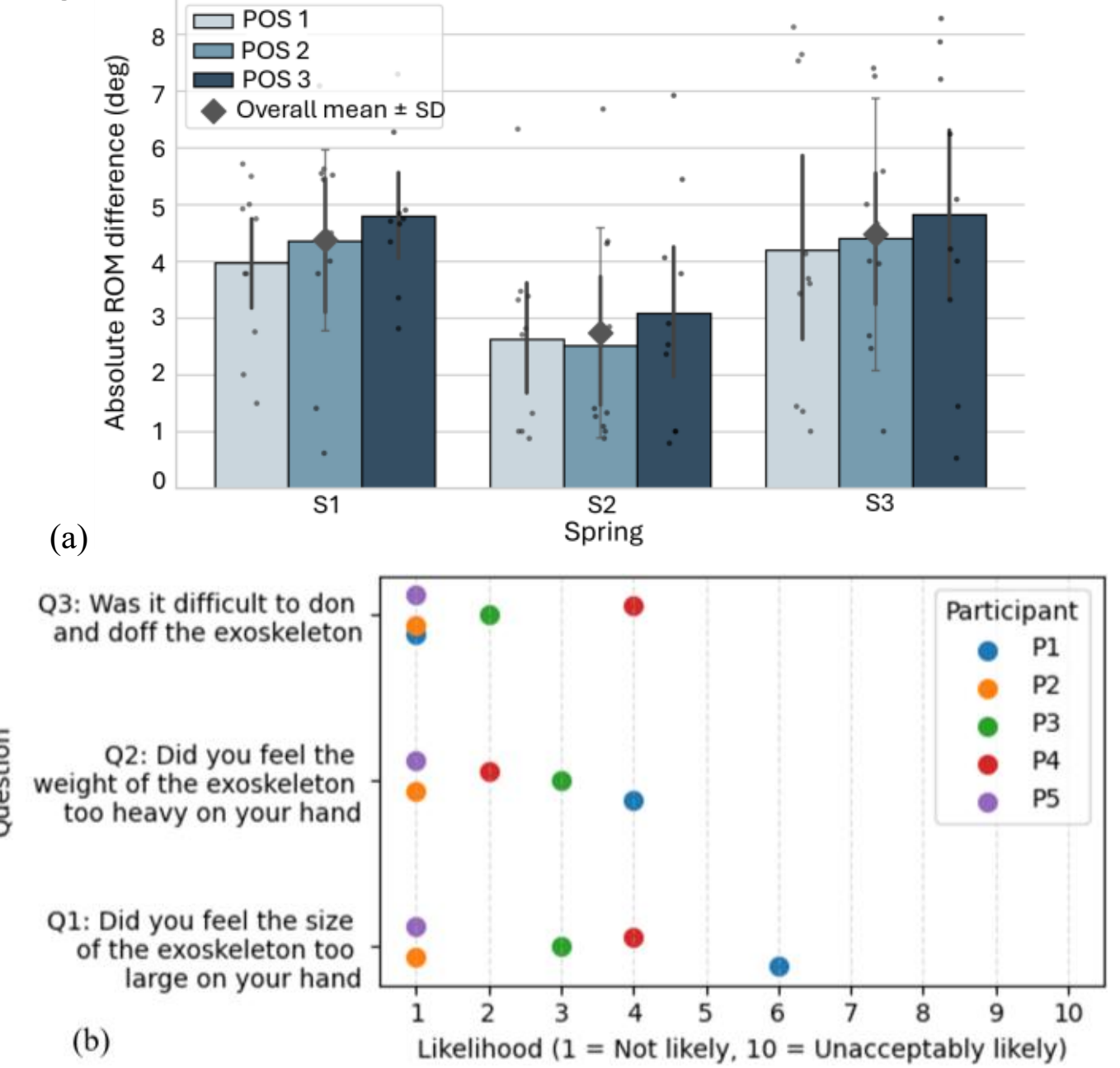

Fig. 8 (a) Repeatability across spring configurations per position ($|T_1 - T_2|$). Bars show mean ± SD; points indicate trials; diamonds denote overall mean ± SD and (b) Participants' Likert scale response per question.

## VII. CONCLUSION

The proposed single-cable, clock-spring-assisted wrist exoskeleton demonstrates a compact and lightweight solution for supporting abduction-adduction motion. By integrating a preloaded clock spring, the system maintains continuous cable tension without relying on antagonistic actuation or active control, effectively reducing slack and peak motor torque demands. This passive assistance mechanism enables smoother bidirectional motion while preserving transparency and control fidelity.

Importantly, the study demonstrates that simulation-derived stiffness and pretension parameters can effectively guide the design of assistive mechanisms, with experimental results showing agreement with model predictions. Combined with lightweight 3D-printed components and streamlined joint architecture, the lightweight and compact design offers enhanced wearability and mechanical simplicity, making it well-suited for portable rehabilitation and assistive applications. Future work should explore pretension adjustment, improved mechanism for ROM limits and extending the simulation-driven design approach to additional joints in a full upper-limb exoskeleton.

## Acknowledgment

This work has been funded by the Aage and Johanne Louis-Hansens Foundation (Grant No. 20-2B- 7273), Denmark at Aalborg University.